\documentclass[journal]{IEEEtran}
\usepackage{blindtext, graphicx}
\usepackage{multirow}
\usepackage{cite}
\usepackage[ruled,vlined,commentsnumbered,linesnumbered,lined,boxed]{algorithm2e}
\usepackage{algorithmic}
\usepackage{amsmath}
\usepackage{xcolor}

\begin{document}

%
\title{SpinalNet: Deep Neural Network with \\Gradual Input}
\author{H M Dipu Kabir, Moloud Abdar, Seyed Mohammad Jafar Jalali, \\Abbas Khosravi, \emph{Senior Member, IEEE}; Amir F Atiya, \emph{Senior Member, IEEE};\\ Saeid Nahavandi, \emph{Fellow, IEEE};  Dipti Srinivasan, \emph{Fellow, IEEE}.

\thanks{H M Dipu Kabir, Moloud Abdar, Seyed Mohammad Jafar Jalali, Abbas Khosravi, and Saeid Nahavandi are with Institute for Intelligent Systems Research and Innovation (IISRI), Deakin University, Australia. (Email: \{hussain.kabir, mabdar, sjalali, abbas.khosravi, saeid.nahavandi\}@deakin.edu.au)}

\thanks{Amir F Atiya in with Cairo University. Email: amir@alumni.caltech.edu}

\thanks{Dipti Srinivasan is with National University of Singapore. Email: dipti@nus.edu.sg}

\thanks{Manuscript received  --, 2020;}
\thanks{accepted-------.}
}


%


\maketitle

\begin{abstract}
Deep neural networks (DNNs) have achieved the state of the art performance in numerous fields. However, DNNs need high computation times, and people always expect better performance in a lower computation. Therefore, we study the human somatosensory system and design a neural network (SpinalNet) to achieve higher accuracy with fewer computations. Hidden layers in traditional NNs receive inputs in the previous layer, apply activation function, and then transfer the outcomes to the next layer. In the proposed SpinalNet, each layer is split into three splits: 1) input split, 2) intermediate split, and 3) output split. Input split of each layer receives a part of the inputs. The intermediate split of each layer receives outputs of the intermediate split of the previous layer and outputs of the input split of the current layer. The number of incoming weights becomes significantly lower than traditional DNNs. The SpinalNet can also be used as the fully connected or classification layer of DNN and supports both traditional learning and transfer learning. We observe significant error reductions with lower computational costs in most of the DNNs. Traditional learning on the VGG-5 network with SpinalNet classification layers provided the state-of-the-art (SOTA) performance on QMNIST, Kuzushiji-MNIST, EMNIST (Letters, Digits, and Balanced) datasets. Traditional learning with ImageNet pre-trained initial weights and SpinalNet classification layers provided the SOTA performance on STL-10, Fruits 360, Bird225, and Caltech-101 datasets. The scripts of the proposed SpinalNet are available at the following link: https://github.com/dipuk0506/SpinalNet

\end{abstract}


%
\IEEEpeerreviewmaketitle

\section{Introduction}

Deep Neural Networks (DNNs) have brought the state of the art performance in various scientific and engineering fields \cite{byerly2020branching, ciregan2012multi, bengio2013representation, kowsari2018rmdl}. DNNs usually have a large number of input features, as the consideration of more parameters usually improves the accuracy of the prediction. The size of the first hidden layer is critical. A small first hidden layer fails to propagate all input features properly, and a large first hidden layer increases the number of weights drastically. Another limitation of the traditional DNNs is the vanishing gradient. When the number of layers is large, the gradient is high at parameters near the output, and the gradient becomes negligible at parameters near inputs. DNN training becomes difficult due to the vanishing gradient problem.

\begin{figure}
  \centering
  \includegraphics[clip, trim=3.6cm .7cm 12cm 1cm, width=3.2in,angle=0]{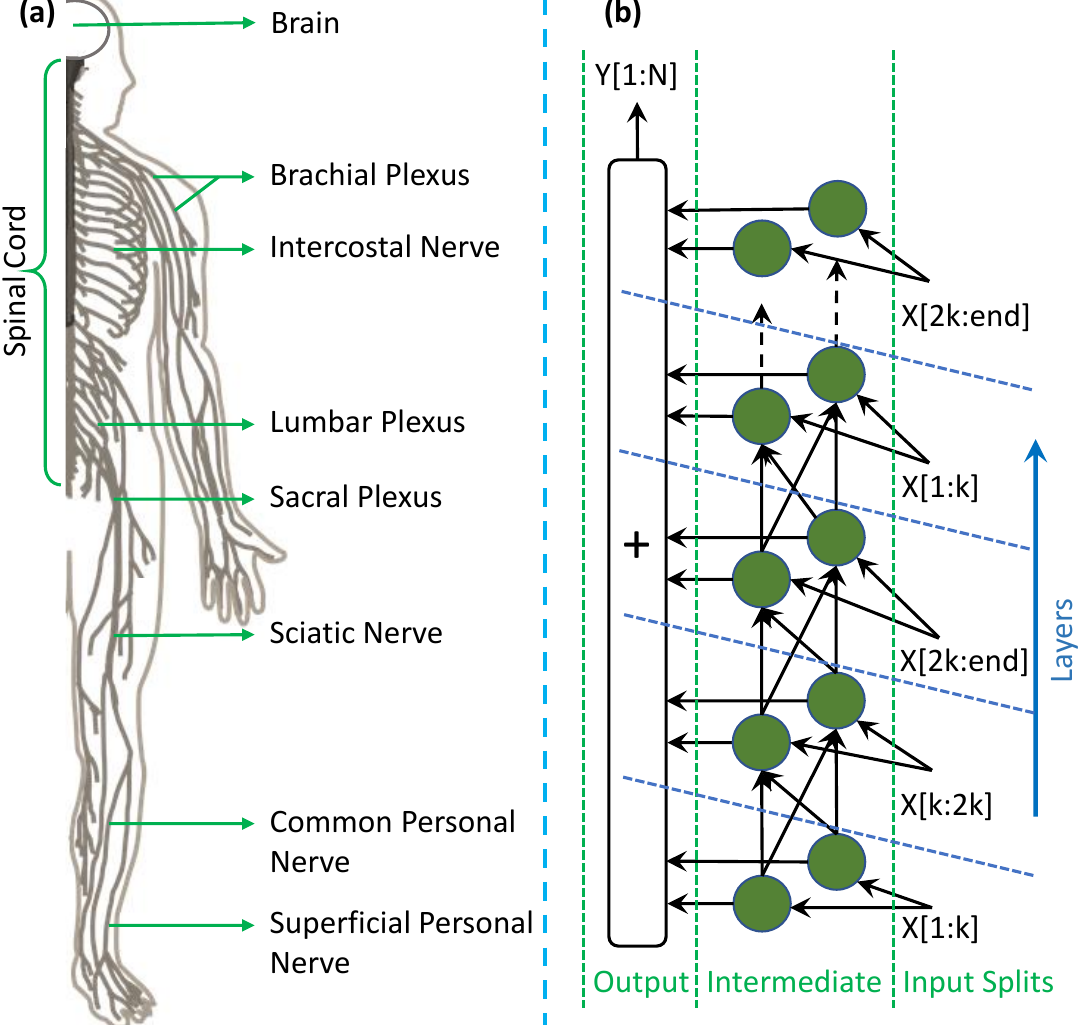}
  \caption{We develop SpinalNet by mimicking several characteristics of the human somatosensory system to receive large input efficiently and to achieve better performance. (a) How our spinal cord is connected to our body for receiving and sending sensory signals from our body. (b) Structure of the proposed SpinalNet. Each layer of the proposed NN is split into input split, intermediate split, and output split. Each intermediate split receives a portion of the input. All intermediate splits except the intermediate split of the first layer also receive outputs of the previous intermediate split. The output spit adds the weighted outputs of all intermediate splits. The user can also construct and train a SpinalNet for any arbitrary number of inputs, intermediate neurons, and outputs.   }
  \label{Human_SN}
\end{figure}

The human brain receives a lot of information from our skin. Numerous tactile sensory neurons send the sense of touch, heat, vibration, etc. from different parts of our body. They can sense pressure, heat, vibrations, complex textures, hardness, state of matter, etc. \cite{okamoto2012psychophysical}. Humans can have different touch sensitivity over time. Although the exact mechanism is unknown to humans, the current knowledge base states a tremendous function of our spinal cord neurons. The human spinal cord receives senses of touch from different locations in different parts of it. Multiple vertebrae can be connected to one internal organ too. Fig. \ref{Human_SN}(a) presents simplified rough connections between human touch-sensors and the spinal cord.  Researchers have developed convolutional neural networks (CNN) by mimicking the functionality of the cats' visual cortex, and that brings a significant improvement in the accuracy of NNs \cite{hubel1962receptive}. The miraculous spinal architecture of humans and the recent success of CNNs motivate us to develop a neural network with gradual inputs.

A well-known approach to reducing computation is pooling \cite{lawrence1997face}. However, pooling causes loss of information. 
Popular solutions to the vanishing gradient problem are ResNet and DenseNet. They allow shortcut connections over different layers. Therefore, the gradient remains high at neurons near the input \cite{he2016deep}. ResNets usually provide better performance with increasing depth and can be as deep as thousands of layers. However, there is a slight marginal improvement in ResNet with increased depth. Moreover, very deep ResNets have a problem of diminishing feature reuse. Therefore, Sergey et al. proposed wide residual networks \cite{zagoruyko2016wide } and achieved superior performance.  Zifeng et al. also received superior performance with shallow and wide NNs \cite{wu2019wider}. Gao et al. proposed DenseNet, where all layers are connected \cite{huang2017densely}. DenseNet training is faster and provides better performance in most situations due to two reasons i) all layers of DenseNet are connected, ii) they made DenseNet narrower than the ResNet. When all layers are connected, the gradient and feature-reuse do not vanish over layers.  However, as all layers are connected, an increment of the network size by one layer needs connections to that layer from all existing layers. Therefore, deep DenseNets are computationally expensive. Adaptive Structural Learning of Neural Networks(AdaNet) performs both connecting neurons and optimizing weights during the training. Consideration of all possible connections of a DNN is computationally intensive. The inauguration of a new neuron requires the consideration of connecting the neuron to other neurons. Therefore, AdaNet is suitable for shallow NNs \cite{cortes2017adanet, kabir2019partial}.

Although DNNs have achieved SOTA performances in numerous fields, DNNs still suffer from large computational overheads during training and execution \cite{jalali2019optimal}. This paper proposes the SpinalNet, presented in Fig. \ref{Human_SN}(b) to improve performance with smaller computational overhead. The proposed structure with gradual and repetitive input capabilities enables NNs to achieve promising results with fewer parameters. We investigate the proposed SpinalNet as the fully connected layer of the VGG-5 network and receive SOTA performance in four MNIST datasets. We apply the transferred initialization with SpinalNet fully connected layers and receive SOTA performance in STL-10, Fruits 360, and Caltech-101 datasets. Subsection IID explains the term: transferred initialization. We investigate SpinalNet and its variations in more than seventeen different datasets. We also receive promising results in other datasets with at least one variant of SpinalNet.

We organize the rest of the paper as follows: Section II presents the theoretical background of SpinalNet. The section discusses the similarity between the human spinal cord and the proposed SpinalNet, proves the universal approximation of the SpinalNet, and discusses transferred initialization. Section III reports the experimental results of SpinalNet with other competitors for solving regression and classification problems, Section IV presents the prospects of SpinalNet, and Section V is the concluding section.

\section{Theoretical Background}
Deep NNs can be of convolutional or non-convolutional type. The structure of non-convolutional NNs consists of fully connected input, hidden, and output layers. Deep convolutional NNs contain convolutional layers and fully connected layers, except the SqueezeNet \cite{iandola2016squeezenet}. SqueezeNet uses global average pooling instead of a fully connected layer. However, SqueezeNet is not one of the best models in terms of accuracy at this moment. DNNs usually have a large number of parameters in the fully connected or classifier layers. Especially DNNs of VGG series has millions of parameters at the fully connected layer. DNNs of ResNet or DenseNet series lack neurons in their fully connected layer. Therefore, there is a linear relationship between features and outputs. However, there can be non-linear relationships between features and outputs. Consideration of linear relationships may result in inferior performances. Therefore we demonstrate Spinal layers instead of the traditional hidden- layers. We demonstrate Spinal layers for both shallow NNs and as the fully-connected part of several popular Deep CNNs. 

\begin{figure}
  \centering
  \includegraphics[width=2.7in,angle=0]{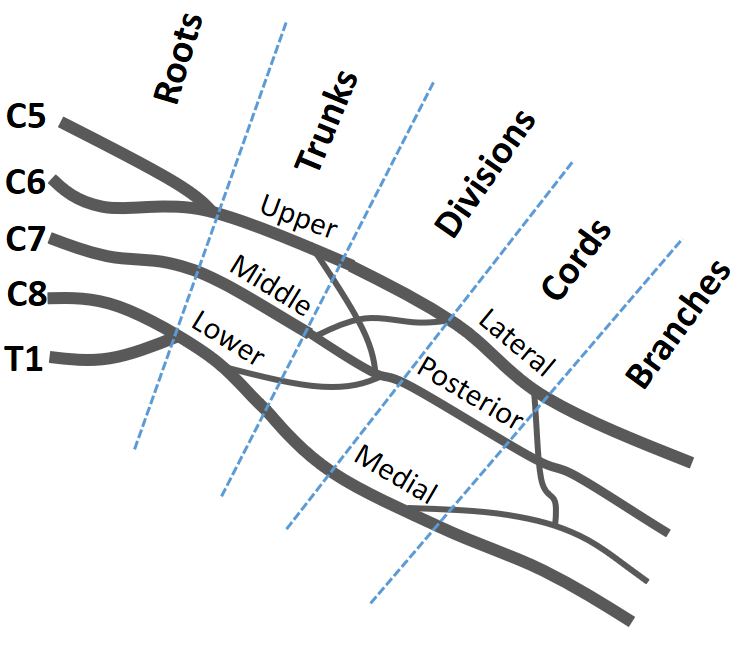}
  \caption{A portion of the human nerve plexus; known as the brachial plexus. The information of any touch or pain reaches the brain through the nerve plexus and the spinal cord. The nerve plexus is a network of intersecting nerves. Our spinal cord receives information gradually. 
  Here, C5-C8 and T1 are vertebrae \cite{chang2019vascularized} in the human skeleton.}
  \label{Hsensory}
\end{figure}

\subsection{Human Somatosensory System and the Spinal Cord}
Although the exact mechanism of the human somatosensory system is not well understood, we find several similarities between the human spinal cord and the proposed neural network \cite{d2008spinal}. We aim to mimic the following characteristics of human spinal cord:
\begin{enumerate}
	\item Gradual input and nerve plexus. 
	\item Voluntary and involuntary actions.
	\item Attention to pain intensity. 
\end{enumerate}

Sensory neurons reach the spinal cord through a complex network, known as the nerve plexus. Fig. \ref{Hsensory} represents a portion of the nerve plexus. A single vertebra does not receive all of the information. The tactile sensory network covers millions of sensors. Furthermore, the human tactile system is more stable compared to the vision or the auditory systems, as there are fewer touch-blind patients, compared to the number of blind patients. The nerve plexus network sends all tactile signals to the spinal cord gradually. Different locations of a spinal cord receive the pain of the leg and the pain of the hand \cite{d2008spinal}.
The neurons existing in the vertebral column are responsible for transferring the sense of touch to the brain and may take some actions. Our brain can control the spinal neurons to increase or decrease the pain intensity \cite{sprenger2012attention}. Sensory neurons may also convey information to the lower motor before getting instruction from the brain. This miraculous procedure is called involuntary or reflex movements.

\subsection{Proposed SpinalNet}

The proposed SpinalNet has the following similarities with the above-mentioned features of the human spinal cord.
\begin{enumerate}
    \item Gradual input 
    \item Local output and probable global influence
    \item Weights reconfigured during training
\end{enumerate}
Similar to our spinal cord, the proposed SpinalNet takes inputs gradually and repetitively. Each layer of the SpinalNet contributes towards the local output (reflex). The SpinalNet also sends a modulated version of inputs towards the global output (brain). The NN training process configures weights based on the training data. Spinal cord neurons are also get configured for tuning the pain sensitivity of different sensories of our body.

Fig. \ref{Human_SN} demonstrates the structure of the proposed SpinalNet. The network structure consists of input sub-layers, intermediate sub-layers, and an output layer. The input is split and sent to the intermediate sub-layers of multiple hidden layers. In Fig. \ref{Human_SN}, the intermediate sub-layers contain two neurons per hidden layer. The number of intermediate neurons can be changed according to the user. However, both the number of intermediate neurons and the number of inputs per layer are regularly kept small to reduce the number of multiplication. As typically the number of inputs and the intermediate hidden neurons per layer allocates a small amount, the network may have an under-fit shape. As a consequence, each layer receives inputs from the previous layer. Since the input is repeated, if one important feature of input does not impact the output in one hidden layer, the feature may impact the output in another hidden layer. The intermediate sub-layers contain a nonlinear activation function and the output layer contains the linear activation function. 
In Fig. \ref{Human_SN}(b), the input values are split into three rows. These rows are assigned to the different hidden layers in a repeated manner. 

\subsection{Universal Approximation of the Proposed SpinalNet}
\begin{figure}
  \centering
  \includegraphics[width=3.5in,angle=0]{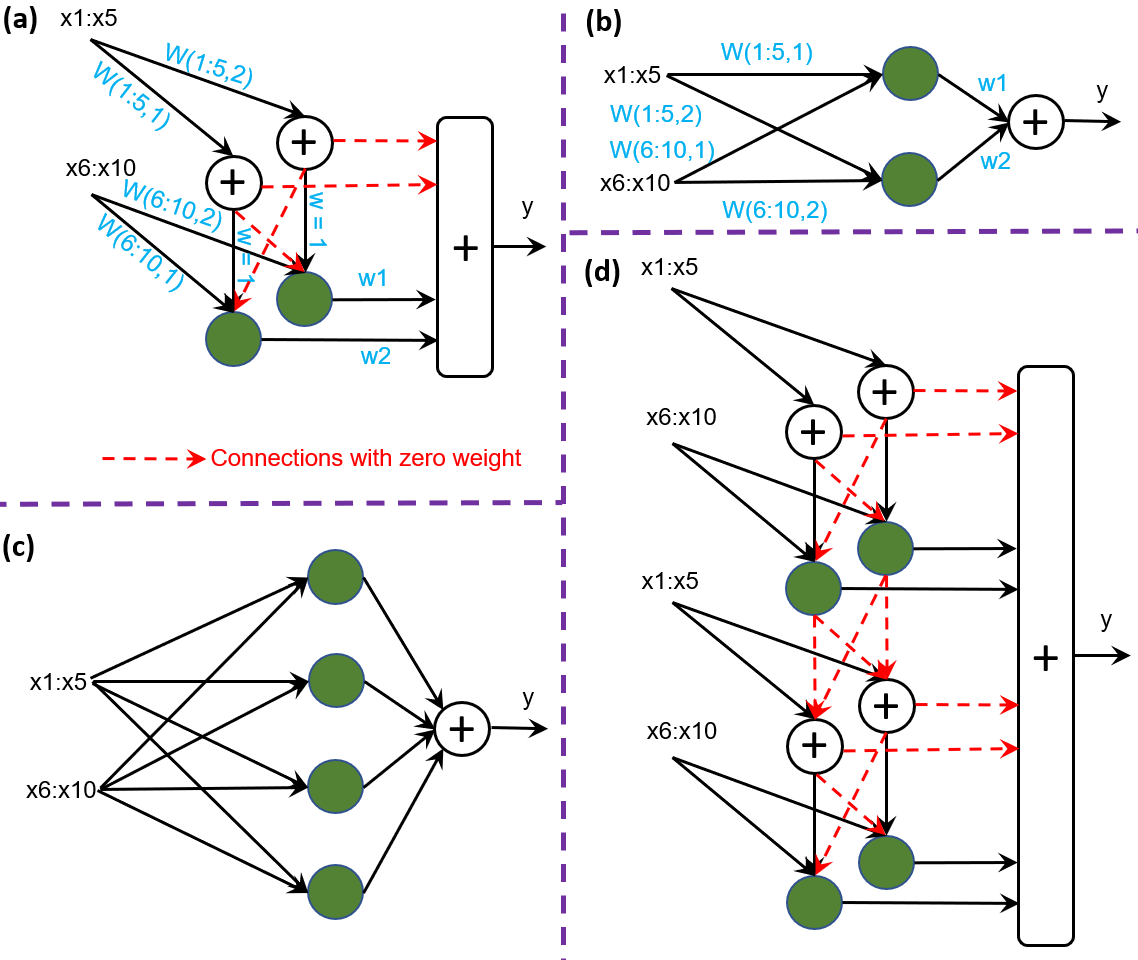}
  \caption{The visual proof of the universal approximation of the SpinalNet. A simplified version of SpinalNet in (a) can act as a NN of a single hidden layer, drawn in (b). Similarly, a 4 layer SpinalNet in (d) can be equivalent to a NN of one hidden layer (HL), containing four neurons, shown in (c). }
  \label{UApprox}
\end{figure}
Proposing a new NN architecture raises a question about its universal approximability \cite{lin2018resnet,fan2020universal}. Therefore, we prove the universal approximation. The traditional mathematical proof of the universal approximation theorem contains scholarly and esoteric equations. However, we aim to make the paper equation-free to attract general audiences. Therefore, we prove the universal approximation with the following approach.
\begin{enumerate}
  \item Single hidden layer of a NN with large width is a universal approximator \cite{csaji2001approximation}.
  \item If we can prove that SpinalNet of a large number of layers is equivalent to the single hidden layer of a large width NN, the universal approximation is proved.
\end{enumerate}

Fig. \ref{UApprox} presents how a simpler version of SpinalNet can be equivalent to a single hidden layer NN. In Fig. \ref{UApprox}(a), a SpinalNet with two hidden layers (HLs), each layer containing two neurons is simplified. The neurons of the first layer contain the purely linear function. Therefore, the first layer gives only the weighted sum of $x_1$ to $x_5$ inputs. Outputs of each hidden neuron for the first hidden layer only go to a similar neuron of the second hidden layer. Cross connections and connections from the first hidden layer to output are disconnected by assigning zero weights. The second hidden layer receives the weighted summation from $x_6$ to $x_{10}$. It also receives the weighted summation from  $x_1$ to $x_5$ of the previous layer. Then, the neurons of this layer apply an activation function to the weighted sum of $x_1$ to $x_{10}$. Therefore, these two layers are equivalent to a neural network of single HL, containing two hidden neurons, shown in Fig. \ref{UApprox}(b). A simplified version of SpinalNet with four HLs, containing two neurons in each layer is shown in Fig. \ref{UApprox}(d). Similarly, the SpinalNet in Fig. \ref{UApprox}(d) is also equivalent to a NN with one HL, containing four neurons in Fig. \ref{UApprox}(c). In the supplementary material section, we have proved that any activation function in the first hidden layer can be approximated as the `purelin' function for a specific combination of weights and biases.

Similarly, a deep SpinalNet can be equivalent to a NN of a single hidden layer, containing a large number of neurons.  A NN with a single hidden layer and a large number of neurons achieves the universal approximation. Therefore, a deep SpinalNet also has universal approximability.

\subsection{Transferred Initialization}
 Transfer of learning is an efficient technique of using previously acquired knowledge and skills in novel problems. It is also similar to educating humans with a much broader syllabus to achieve competencies for an unpredictable future. One of the most efficient ideas of machine learning is transfer learning(TL), which is similar to the transfer of learning in humans \cite{wolf2019transfertransfo}. DNNs require adequate training samples for proper training. Insufficient training samples may result in poor performance. Transfer learning is an efficient DNN training technique where initial layers of DNN are pre-trained with a large dataset. The corresponding train dataset trains only a few final layers. As a result, the user can get a well-trained NN for the specific dataset of a small sample number, with lower computational overhead \cite{shao2018starcraft}.  The TL is gaining huge popularity these days due to exceptional performance and usability. Many researchers expect TL as the next driver of the commercial success of machine learning \cite{ng2016nuts}. Several standard datasets contain insufficient training samples. CIFAR-100, Caltech-101, STL-10, etc. datasets are examples of such datasets. The traditional DNN training trains the entire DNN. According to the reported results, current traditional DNN training techniques haven't achieved more than 90\% accuracy on the CIFAR-100 dataset but TL has achieved more than 93\% accuracy. 

We apply the TL method without freezing weights of convolutional layers. Therefore, the initial weights are transferred, and the training is traditional. When initial weights are frozen, the optimization happens at a local minimum. While training NNs multiple times without freezing weights, one optimization can potentially happen in the global minima. Moreover, pre-trained initial layers can potentially fail to propagate several important features of the target dataset. That propagation error may result in inferior performance in transfer learning than the transferred initialization.

\section{Results}
In this section, we aim to verify the effectiveness of SpinalNet for regression and classification problems. We apply different variants of the SpinalNet and receive SOTA performance in several datasets. The training procedure follows stochastic gradient descent (SGD) or Adam training technique. We upload training scripts to GitHub to help future researchers.

\subsection{Regression Dataset}
Regression is a less popular topic among the researchers of NN compared to classification. There exist a large number of datasets and organized competition among various algorithms for the classification problem. Regregression lacks popular standard datasets and competitions. Therefore, we compare our SpinalNet with the PyTorch regression example, developed by Ben Phillips \cite{Regression}. The example considers a single input and a single output. Following the example, we apply the Adam algorithm \cite{kingma2014adam} for the optimization purpose. The loss function used in the experiments is the mean square error (MSE), the learning rate is equal to 0.01, and the number of the epoch is equal to 200. We changed the problem to eight variables and tried different combinations of variables with the same level of noise. Combinations are 1) summation of variables ($\sum x$), 2) sine of summation of variables ($sin(\sum x)$), and 3) product of variables ($\prod x$), and 4) sine of product of variables ($sin(\prod x)$).  We record the MSE at 100 and 200 epochs. The default code \cite{Regression} shows the MSE of the last epoch, but our code shows the minimum MSE among the current and previous epochs. We segment inputs into two groups, each containing four inputs.  

The number of hidden neurons in traditional NN is 300 \cite{Regression}. The number of hidden neurons in SpinalNet is also 300. The number of multiplication in traditional NN is 21.7k, and the number of multiplications in SpinalNet is 14k. The SpinalNet achieves a 35.5\% reduction in the number of multiplications. Memory requirements and execution times are two important determinants of the quality of the neural network besides the accuracy or error value. Recent papers on the proposal of novel NN are presenting the number of parameters as an indication of memory and time requirement \cite{touvron2020fixing, du2020spinenet}. The number of parameters in the traditional NN is 22k, and the number of parameters in SpinalNet is 14.3k. 
There are eight combinations for MSE comparisons, as shown in Table \ref{Regression1}. Superior performances are highlighted as bold characters. The SpinalNet performs better in six out of eight combinations.

\begin{table}
\centering
\caption{Comparison between Traditional Feed-forward NN and SpinalNet for Regression datasets. }
\label{Regression1}
\begin{tabular}{|c|c|c|c|}
\hline
 Neural Network & Data & \multicolumn{2}{|c|}{MSE ($10^{-3}$ Unit)}    \\ \cline{3-4}
                &      & 100 Epoch & 200 Epoch    \\ \hline 
 Feed-forward NN & 8 Var. $\sum x$ & 1.178&0.887    \\ \cline{2-4} 
 Two Hidden Layers& 8 Var. $sin(\sum x)$ &1.918&\bf{1.086}    \\ \cline{2-4}  
 200, 100 Neurons & 8 Var. $\prod x$ & \bf{3.875} &3.875    \\ \cline{2-4} 
 \cite{Regression} & 8 Var. $sin(\prod x)$ &3.403 &1.554    \\ \hline 
 
 SpinalNet & 8 Var. $\sum x$ &\bf{1.007}& \bf{0.855}    \\ \cline{2-4} 
 6 Hidden Layers& 8 Var. $sin(\sum x)$ & \bf{1.912}&1.219    \\ \cline{2-4}  
 50 Neurons Each Layer & 8 Var. $\prod x$ &3.966 &\bf{2.217}    \\ \cline{2-4} 
 Half Input Each Layer & 8 Var. $sin(\prod x)$ & \bf{0.910} &\bf{0.910}    \\ \hline 
 
\end{tabular}
\end{table}

\subsection{Classification: Learning from Random Initialization}
We train several existing networks and different variations of SpinalNet with random initialization on MNIST, Fashion-MNIST, KMNIST, QMNIST, EMNIST, CIFAR-10, and CIFAR-100  image classification datasets. We train each model ten times and present average and best results.

\subsubsection{MNIST}
The MNIST dataset is one of the most popular datasets for investigating image classification algorithms due to its simplicity and small size. We compare our SpinalNet with PyTorch CNN \cite{MNIST_CNN}. A hidden layer of fifty neurons joins them with the output. The default CNN code provides 98.17\% accuracy. We investigate the same NN with a SpinalNet fully connected (FC) layer. The FC layer consists of six sub-hidden layers, each layer containing eight neurons. CNN with Spinal FC provides 98.44\% accuracy. That structure brings more than a 48.5\% reduction in multiplication and a 4\% reduction in the activation functions on the fully connected layer. The first segment in Table \ref{Perform_tab} presents results on the MNIST dataset. The SpinalNet reduces the overall number of parameters and increases performance significantly.

As VGG models perform very well with the MNIST datasets, we incorporate the SpinalNet with the VGG-5 network \cite{VGG5}. VGG-5 with the Spinal fully connected layer provides a near state-of-the-art performance. We receive 99.72\% accuracy with VGG-5 (Spinal FC). According to our literature search, it is one of the top twenty reported accuracies. We perform the random rotation of 10 degrees and the random perspective PyTorch library functions to enhance data.

\subsubsection{Fashion-MNIST}
The Fashion-MNIST data is quite similar to MNIST. It also contains 28 $\times$ 28 grayscale images and the output contains 10 classes. Therefore, MNIST codes are executable to the Fashion-MNIST data without any modification. The same NN is applied to compute CNN and CNN (Spinal FC).  

We receive 94.68\% accuracy with VGG-5 (Spinal FC). The default VGG-5 provides 94.63\% accuracy. Therefore, the Spinal FC provides better performance with a lower number of multiplications. We apply random rotation and random resized crops to enhance the data. According to our literature search, it is one of the top five reported results.

\subsubsection{Kuzushiji-MNIST}
Kuzushiji-MNIST or shortly, KMNIST is a Japanese character recognition dataset. The data format of KMNIST is also the same as the MNIST data, and the same codes can be applied. Table \ref{Perform_tab} presents the performance of codes. We receive superior performance by using the Spinal FC layer with CNN.

We receive 99.15\% accuracy with VGG-5 (Spinal FC). 
It is a new SOTA for the Kuzushiji-MNIST dataset \cite{nokland2019training,tissera2019context}. We apply the random perspective and the random rotation to enhance the data.
The data augmentation for VGG-5 and VGG-5  (Spinal FC) are the same. The default VGG-5 provides 98.94\% accuracy. Therefore, the Spinal FC provides better performance with a lower number of multiplications.

\subsubsection{QMNIST}
QMNIST is a recently published English digit recognition dataset. The QMNIST dataset contains fifty thousand test images. The dimensions of inputs and outputs are the same as the dimension of inputs and outputs of the MNIST dataset. Therefore, the same code is executable for the QMNIST dataset. The fourth segment of table \ref{Perform_tab} presents QMNIST results. The default PyTorch CNN provides 97.82\% accuracy on the QMNIST data. CNN(Spinal FC) provides 97.97\% and 98.07\% accuracy for the spinal-layer size of eight and ten respectively. 

The VGG-5 receives 99.66\% accuracy. The VGG-5 (Spinal FC) receives 99.68\% accuracy. According to our literature search, we have received SOTA performance for the QMNIST dataset. We apply the random perspective and random rotation functions to achieve these results.

\subsubsection{EMNIST}
The EMNIST dataset contains several hand-written character datasets. These datasets are derived from the NIST database and converted to a 28$\times$28 pixel grey-scale image format. The EMNIST(digits) dataset also has ten classes. The accuracy of NNs on the EMNIST (digit) data are available in the fifth segment of table \ref{Perform_tab}. The default PyTorch CNN provides 98.89\% accuracy. 
The VGG-5 provides 99.81\% accuracy and VGG-5 (Spinal FC) provides 99.82\% accuracy. According to our literature search, we have received SOTA performance for the EMNIST (digits) dataset.

The default PyTorch CNN provides 87.57\% accuracy for EMNIST (letters) dataset. 
The VGG-5 provides 95.86\% accuracy and VGG-5 (Spinal FC) provides 95.88\% accuracy. According to our literature search, we have received SOTA performance for the EMNIST (letters) dataset.

The default PyTorch CNN provides 79.61\% accuracy for EMNIST (balanced) dataset. 
The VGG-5 provides 91.04\% accuracy. VGG-5 (Spinal FC) provides 91.05\% accuracy. According to our literature search, we have received SOTA performance for the EMNIST (balanced) dataset. We apply the random perspective and the random rotation functions to enhance all EMNIST datasets.

\subsubsection{CIFAR-10 Dataset}
The CIFAR-10 dataset is one of the most popular datasets in computer vision \cite{huang2019gpipe}. The CIFAR-10 dataset contains 32 $\times$ 32 sized color images. There are ten output classes. The dataset contains fifty thousand images for training and ten thousand images for testing. 

We train and report results with ResNet-18 \cite{he2016deep} and VGG-19\cite{simonyan2014very} models. Although results do not improve with the ResNet-18, we observe superior results with the VGG-19 model. Higher accuracy is achieved with a lower number of parameters with the VGG-19 model. 8$^{th}$ segment of Table \ref{Perform_tab} presents results of traditional training on CIFAR-10 dataset.

\begin{table*}
\centering
\caption{Performance of the SpinalNet and several popular Nets on Different Classification Datasets}
\label{Perform_tab}
\begin{tabular}{|c|c|c|c|c|c|c|c|}
\hline
 Data  & Model & Size of Fully Connected Layer & Epoch & \multicolumn{2}{|c|}{Test Accuracy} & Error Reduction & Parameters  \\ \cline{5-6}
 &&&&Average&Best&(Best)&
 \\\hline 

      & CNN\cite{MNIST_CNN}& 1HL, 50 Neurons &8 & 97.98\% &98.17\%&- & 21.84k \\ \cline{2-8}
 MNIST  & CNN (Spinal FC) & 6HL, 8 Neurons Per Layer &8 &98.32\%&98.44\%  & 14.8\% &13.82k\\ \cline{2-8}
 \cite{deng2012mnist} & CNN(Spinal FC) & 6HL, 10 Neurons Per Layer &8 &98.37\%&98.48\% &16.9\%&16.05k\\ \cline{2-8}
      
      & VGG-5\cite{VGG5} & 1HL, 512 Neurons &100&99.69\%&99.72\% &- &3.646M\\ \cline{2-8}
      & VGG-5 (Spinal FC) & 4HL, 128 Neurons Per Layer &100&99.69\%&99.72\% &0.0\% &3.630M\\ \hline \hline 
      
        & CNN\cite{MNIST_CNN}& 1HL, 50 Neurons &8 &83.59\%&84.10\%&-& 21.84k \\ \cline{2-8}
 Fashion-MNIST  & CNN (Spinal FC) & 6HL, 8 Neurons Per Layer  &8 &85.31\%&85.98\%  & 11.8\%&13.82k\\ \cline{2-8}
  \cite{xiao2017fashion}    & CNN(Spinal FC) & 6HL, 10 Neurons Per Layer &8 &85.71\%&86.61\% &15.8\%&16.05k\\ \cline{2-8}
     
      & VGG-5\cite{VGG5} & 1HL, 512 Neurons &100&94.02\%&94.63\% &- &3.646M\\ \cline{2-8}
      & VGG-5 (Spinal FC) & 4HL, 128 Neurons Per Layer &100&94.19\%&94.68\% &0.9\% &3.630M\\ \hline  \hline 
      
        & CNN\cite{MNIST_CNN}& 1HL, 50 Neurons &8 &83.24\%&84.48\%&- & 21.84k\\ \cline{2-8}
 Kuzushiji-MNIST  & CNN (Spinal FC) & 6HL, 8 Neurons Per Layer &8 &85.98\%&87.94\%  & 22.3\%&13.82k\\ \cline{2-8}
     (KMNIST) \cite{clanuwat2018deep} & CNN (Spinal FC) & 6HL, 10 Neurons Per Layer &8 &86.48\%&88.25\% &24.3\%&16.05k\\ \cline{2-8}
      
      & VGG-5\cite{VGG5} & 1HL, 512 Neurons &200 &98.76\%&98.94\%&- &3.646M\\ \cline{2-8}
      & VGG-5 (Spinal FC) & 4HL, 128 Neurons Per Layer &200 &99.05\%&99.15\%&19.8\% &3.630M\\ \hline \hline

        & CNN\cite{MNIST_CNN}& 1HL, 50 Neurons &8 &97.66\%&97.82\%&-& 21.84k \\ \cline{2-8}
 QMNIST  & CNN (Spinal FC) & 6HL, 8 Neurons Per Layer &8 &97.81\%&97.97\%  & 6.9\%&13.82k\\ \cline{2-8}
   \cite{yadav2019cold}   & CNN(Spinal FC) & 6HL, 10 Neurons Per Layer &8 &98.00\%&98.07\% &11.5\%&16.05k\\ \cline{2-8}
     
      & VGG-5\cite{VGG5} & 1HL, 512 Neurons &100 &99.63\%&99.66\%&- &3.646M\\ \cline{2-8}
      & VGG-5 (Spinal FC) & 4HL, 128 Neurons Per Layer &100 &99.64\%&99.68\%&5.9\% &3.630M\\ \hline \hline 
      
              & CNN\cite{MNIST_CNN}& 1HL, 50 Neurons &8 &98.80\%&98.89\%&- & 21.84k\\ \cline{2-8}
 EMNIST  & CNN (Spinal FC) & 6HL, 8 Neurons Per Layer &8 &99.03\%&99.12\%  & 20.7\%&13.82k\\ \cline{2-8}
  (Digits) \cite{cohen2017emnist}   & CNN(Spinal FC) & 6HL, 10 Neurons Per Layer &8 &99.07\%&99.16\% &24.3\%&16.05k\\ \cline{2-8}
      
      & VGG-5\cite{VGG5} & 1HL, 512 Neurons &50 &99.75\%&99.81\%&- &3.646M\\ \cline{2-8}
      & VGG-5 (Spinal FC) & 4HL, 128 Neurons Per Layer & 50 &99.75\%&99.82\% & 5.3\% &3.630M\\ \hline \hline 
      
    & CNN\cite{MNIST_CNN}& 1HL, 50 Neurons &8 &87.29\%&87.57\%&- & 21.84k\\ \cline{2-8}
 EMNIST  & CNN (Spinal FC) & 6HL, 8 Neurons Per Layer &8 &89.88\%&90.07\%  & 20.11\%&13.82k\\ \cline{2-8}
  (Letters) \cite{cohen2017emnist}   & CNN(Spinal FC) & 6HL, 10 Neurons Per Layer &8 &90.02\%&90.23\% &21.4\%&16.05k\\ \cline{2-8}
     
      & VGG-5\cite{VGG5} & 1HL, 512 Neurons &200 &95.71\%&95.86\%&- &3.646M\\ \cline{2-8}
      & VGG-5 (Spinal FC) & 4HL, 128 Neurons Per Layer & 200 &95.79\%&95.88\% & 0.5\% &3.630M\\ \hline \hline 
      
    & CNN\cite{MNIST_CNN}& 1HL, 50 Neurons &8 &78.99\%&79.61\%&- & 21.84k\\ \cline{2-8}
 EMNIST  & CNN (Spinal FC) & 6HL, 8 Neurons Per Layer &8 &82.13\%&82.77\%  & 15.50\%&13.82k\\ \cline{2-8}
  (Balanced) \cite{cohen2017emnist}   & CNN(Spinal FC) & 6HL, 10 Neurons Per Layer &8 &82.57\%&83.21\% &17.66\% &16.05k\\ \cline{2-8}
      
      & VGG-5\cite{VGG5} & 1HL, 512 Neurons &200 &90.71\%&91.04\%&- &3.646M\\ \cline{2-8}
      & VGG-5 (Spinal FC) & 4HL, 128 Neurons Per Layer & 200 &90.73\%&91.05\% & 0.1\% &3.630M\\ \hline \hline

   & ResNet-18 \cite{he2016deep}&Only Output Layer&150 &90.65\%&91.98\%&- &3.14M\\ \cline{2-8}
CIFAR-10  & ResNet-18 (Spinal FC) & 4HL, 20 Neurons Per Layer &150 &90.39\%&91.42\%  & -7.0\% &3.22M\\ \cline{2-8}
    \cite{krizhevsky2009learning}   & VGG-19\cite{simonyan2014very} & 2HL, 4096 Neurons Each &200 &90.48\%&90.75\% &-&38.96M\\ \cline{2-8} 
    & VGG-19 (Spinal FC) &4HL, 512 Neurons Each &200 &90.97\%&91.40\%&7.0\% & 20.16M\\ \hline \hline

      
  & ResNet-18 \cite{he2016deep}&Only Output Layer&30 &64.41\%&65.04\%&- &3.16M\\ \cline{2-8}
   CIFAR-100 & ResNet-18 (Spinal FC) & 4HL, 128 Neurons Per Layer &30 &62.93\%&63.60\%  & -4.1\% &4.66M\\ \cline{2-8}

      \cite{krizhevsky2009learning}& VGG-19\cite{simonyan2014very} & 2HL, 4096 Neurons Each &150 &61.69\%&62.05\% &- &39.33M\\ \cline{2-8}
      & VGG-19 (Spinal FC) &4HL, 512 Neurons Each &150 &64.12\%&64.77\%&7.2\% &20.20M \\ \hline 

\end{tabular}

\end{table*}

\subsubsection{CIFAR-100 Dataset}
The CIFAR-100 is another popular dataset in computer vision \cite{huang2019gpipe}. The CIFAR-100 dataset contains 32 $\times$ 32 sized color images. There are hundred output classes. The dataset contains five hundred images for training and one hundred images for testing for each class. 

We train and report results with ResNet-18 \cite{he2016deep} and VGG-19\cite{simonyan2014very} models. We observe superior results with the VGG-19 model. Higher accuracy is achieved with a lower number of parameters with the VGG-19 model.
The number of neurons in the fully connected layer is reduced to half with the Spinal FC layer. Moreover, the number of multiplication in the FC layer is reduced to 7\%. The VGG-19 has two hidden layers of size 4096. Replacing them with four spinal layers of 512 size reduces the number of multiplications significantly.
 The last segment of Table \ref{Perform_tab} presents results of traditional training with random initialization on CIFAR-100 dataset.
 
However, the Spinal fully connected layer does not improve the performance of ResNet. Although the number of hidden neurons in the fully connected layer is increasing for ResNet, we are getting a lower performance. The probable reason can be a decrease in the gradient in the initial layers of ResNet due to additional layers in the SpinalNet.

\begin{table*}
\centering
\caption{transferred initialization (TI) Performance of the SpinalNet and several popular Nets on Different Classification Datasets}
\label{Perf_TL}
\begin{tabular}{|c|c|c|c|c|c|c|c|}
\hline
 Data  & Model & Size of Fully Connected Layer & Epoch & \multicolumn{2}{|c|}{Test Accuracy} & Error Reduction & Parameters  \\ \cline{5-6}
 &&&&Average&Best&(Best)&
 \\\hline 
 
   & VGG-19\_bn & 2 HL, 4096 Neurons Per Layer & 25  &94.88\%&95.91\%  & - &263.27M\\  \cline{2-8}
  CIFAR-10    & VGG-19\_bn (Spinal FC) &  4HL, 1024 Neurons Per Layer & 25 &95.18\%& 96.00\%  & 2.2\% &198.26M\\  \cline{2-8}
 \cite{krizhevsky2009learning} & Wide\_ResNet-101\_2 & 0 Neurons & 50 &97.43\%& 98.22\%  & - &124.86M\\  \cline{2-8}
 & Wide\_ResNet-101\_2 (Spinal FC) & 4HL, 20 Neurons Per Layer &50 &97.61\%&98.12\% & -5.6\% &124.92M\\ \hline  \hline

   & VGG-19\_bn & 2 HL, 4096 Neurons Per Layer & 25 & 77.82\%& 79.22\%  & - &263.63M\\  \cline{2-8}
 CIFAR-100     & VGG-19\_bn (Spinal FC) &  4HL, 1024 Neurons Per Layer & 25 & 77.93\%& 79.56\%  & 1.6\% &198.63M\\  \cline{2-8}
  \cite{krizhevsky2009learning} & Wide\_ResNet-101\_2 & 0 Neurons & 50 &  86.79\%&87.15\%  & - &125.04M\\  \cline{2-8}
 & Wide\_ResNet-101\_2 (Spinal FC) &  4HL, 512 Neurons Per Layer &50 & 88.00\%&88.34\% & 9.26\% &132.59M\\ \hline \hline

   & VGG-19\_bn & 2 HL, 4096 Neurons Per Layer & 10 & 91.31\%& 92.98\%  & - &263.64M\\  \cline{2-8}
  Caltech-101     & VGG-19\_bn (Spinal FC) &  4HL, 1024 Neurons Per Layer & 10 & 91.92\%& 93.16\%  & 2.6\% &198.63M\\  \cline{2-8}
 \cite{li2004caltech} & Wide\_ResNet-101\_2 & 0 Neurons & 10 & 96.68\%& 97.11\%  & - &125.05M\\  \cline{2-8}
 & Wide\_ResNet-101\_2 (Spinal FC) & 4HL, 101 Neurons Per Layer &10 & 97.04\%&97.32\% & 7.27\% &132.60M\\ \hline \hline

  & VGG-19\_bn & 2 HL, 4096 Neurons Per Layer & 25 & 98.04\%& 98.67\%  & - &264.15M\\  \cline{2-8}
 Bird225 & VGG-19\_bn (Spinal FC) & 4HL, 1024 Neurons Per Layer &25 & 98.62\% &99.02\% & 26.3\% &199.14M\\ \cline{2-8}
 \cite{bird225} & Wide\_ResNet-101\_2 & 0 Neurons & 25 & 98.95\%& 99.38\%  & - &125.30M\\  \cline{2-8} 
 & Wide\_ResNet-101\_2 (Spinal FC) & 4HL, 225 Neurons Per Layer &25 & 99.11\%&99.56\% & 11.1\% &126.11M\\ \hline \hline

  & VGG-19\_bn & 2 HL, 4096 Neurons Per Layer & 25 & 86.42\%&87.21\%  & - &264.03M\\  \cline{2-8}
 Stanford Cars & VGG-19\_bn (Spinal FC) &  4HL, 1024 Neurons Per Layer &25 & 88.06\%&88.72\% & 13.4\% &153.78M\\ \cline{2-8}
 \cite{Stcar} & Wide\_ResNet-101\_2 & 0 Neurons & 25 & 92.89\%& 93.35\%  & - &125.24M\\  \cline{2-8}
 & Wide\_ResNet-101\_2 (Spinal FC) &  4HL, 196 Neurons Per Layer &25 & 92.86\%&93.35\% & 0\% &132.98M\\ \hline \hline

  & VGG-19\_bn & 2 HL, 4096 Neurons Per Layer & 25 & 93.96\%& 94.80\%  & - & 263.27M\\  \cline{2-8}
 SVHN & VGG-19\_bn (Spinal FC) &  4HL, 1024 Neurons Per Layer &25 & 94.61\%&95.26\% & 18.28\% &198.26M\\ \cline{2-8}
 \cite{SVHN_data} & Wide\_ResNet-101\_2 & 0 Neurons & 25 & 97.01\%& 97.80\%  & - &124.86M\\  \cline{2-8}
 & Wide\_ResNet-101\_2 (Spinal FC) & 4HL, 20 Neurons Per Layer &25 & 97.00\%&97.87\% & 3.18\% &124.92M\\ \hline \hline

   & VGG-19\_bn & 2 HL, 4096 Neurons Per Layer & 25 & 89.88\%& 90.28\%  & - & 263.27M\\  \cline{2-8}
  CINIC-10    & VGG-19\_bn (Spinal FC) &  4HL, 1024 Neurons Per Layer & 25 & 90.69\%& 91.00\%  & 8.00\% &198.26M\\  \cline{2-8}
 \cite{darlow2018cinic} & Wide\_ResNet-101\_2 & 0 Neurons & 50 & 91.72\%& 92.15\%  & - &124.86M\\  \cline{2-8}
 & Wide\_ResNet-101\_2 (Spinal FC) & 4HL, 20 Neurons Per Layer &50 & 93.02\%& 93.60\% & 18.47\% &124.92M\\ \hline \hline

   & VGG-19\_bn & 2 HL, 4096 Neurons Per Layer & 10 & 94.97\%& 95.44\%  & - & 263.27M\\  \cline{2-8}
  STL-10    & VGG-19\_bn (Spinal FC) &  4HL, 1024 Neurons Per Layer & 10 & 95.03\%& 95.57\%  & 2.9\% &198.26M\\  \cline{2-8}
 \cite{coates2011analysis} & Wide\_ResNet-101\_2 & 0 Neurons & 10 & 97.83\%& 98.40\%  & - &124.86M\\  \cline{2-8}
 & Wide\_ResNet-101\_2 (Spinal FC) & 4HL, 20 Neurons Per Layer &10 & 98.23\%&98.66\% & 16.3\% &124.92M\\ \hline \hline

   & VGG-19\_bn & 2 HL, 4096 Neurons Per Layer & 25 & 94.65\%& 95.20\%  & - &263.64M\\  \cline{2-8}
  Oxford 102    & VGG-19\_bn (Spinal FC) & 4HL, 1024 Neurons Per Layer & 25 & 94.98\%& 95.46\%  & 32.5\% &198.63M\\  \cline{2-8}
 Flower \cite{nilsback2008automated} & Wide\_ResNet-101\_2 & 0 Neurons & 50 & 98.91\%& 99.39\%  & - & 125.05M\\  \cline{2-8}
 & Wide\_ResNet-101\_2 (Spinal FC) &  4HL, 101 Neurons Per Layer &50 & 99.07\%&99.30\% & -14.7\% &125.32M\\ \hline \hline 

   & VGG-19\_bn & 2 HL, 4096 Neurons Per Layer & 10 & 99.90\% & 99.92\% & - &263.76M\\  \cline{2-8}
  Fruits 360    & VGG-19\_bn (Spinal FC) & 4HL, 1024 Neurons Per Layer & 10 & 99.95\%& 99.96\%  & 60\% &198.75M\\  \cline{2-8}
  \cite{Fruits360} & Wide\_ResNet-101\_2 & 0 Neurons & 10 & 99.94\%&99.96\%  & - & 125.11M\\  \cline{2-8}
 & Wide\_ResNet-101\_2 (Spinal FC) &  4HL, 131 Neurons Per Layer &10 & 99.98\%&100\% & 100\% &125.50M\\ \hline

\end{tabular}

\end{table*}

\begin{table}
\centering
\caption{SOTA Performances of Investigated Datasets in June 2020*}
\label{SOTA_TAB}
\begin{tabular}{|l|l|c|}
\hline
 Data&Model Name  & Accuracy    \\ \hline 
 MNIST&Branching$/$Merging CNN  \cite{byerly2020branching} & 99.84\%   \\ \hline 
 
 Fashion-MNIST&PreAct-ResNet18 + FMix  \cite{harris2020understanding} & 96.36\%   \\ \hline 
 Kuzushiji-MNIST**&	CAMNet3 \cite{tissera2019context}   & 99.05\%    \\ \hline 
 QMNIST**&Deep Regularization  \cite{yoo2020deep} & 99.67\%    \\ \hline 
 EMNIST (Digits)** & DWT-DCT with KNN  \cite{ghadekar2018handwritten} & 97.74\%    \\ \hline 
 EMNIST (Letters)**& TextCaps  \cite{jayasundara2019textcaps} & 95.39\%    \\ \hline 
 EMNIST (Balanced)**& TextCaps  \cite{jayasundara2019textcaps} & 90.46\%    \\ \hline 
 CIFAR-10& 	BiT-L (ResNet) \cite{kolesnikov2019large}  & 99.00\%  \\ \hline 
  CIFAR-100& BiT-L (ResNet) \cite{kolesnikov2019large} & 93.51\%  \\ \hline 
  Caltech-101**& UL-Hopfield \cite{liu2018unsupervised} & 91.00\%  \\ \hline 
  Stanford Cars& DAT \cite{Ngiam2018unsupervised} & 96.20\%  \\ \hline 
  Oxford 102 Flowers& BiT-L (ResNet) \cite{kolesnikov2019large} & 99.63\%  \\ \hline
  STL-10** & NAT-M4 \cite{lu2020neural} & 97.90\% \\ \hline
  CINIC-10 & NAT-M4 \cite{lu2020neural} & 94.80\% \\ \hline
  SVHN & WideResNet-28-10 \cite{Cubuk2019randaugment}  & 99.00\% \\ \hline
  225 Bird Species** & Vgg-16 \cite{bird_vgg16} & 99.10\% \\ \hline
  Fruits 360** & EfficientNet-B1  \cite{duong2020automated} & 100\% \\ \hline
\end{tabular}
\\ $*$ According to the following website and our literature search:  \emph{www.paperswithcode.com/task/image-classification}  \\  
$**$ After the online appearance of the current paper, the performance of SpinalNet will be SOTA performance; unless someone reports a better performance earlier.

\end{table}

\subsection{Classification: Learning from Transferred Initialization}
While learning from scratch, we observe that, although the number of hidden neurons in the fully connected layer is increasing, the ResNet with Spinal FC is getting a lower performance compared to the similar ResNet. The probable reason can be a decrease in the gradient in the initial layers of ResNet due to additional layers. Therefore, we apply the transferred initialization technique to the pre-trained ResNet and VGG networks. We download pre-trained VGG-19\_bn and Wide-Resnet-101 models from Torchvision. These models are pre-trained on the Imagenet dataset. We apply these two models to the following datasets. According to our simulations, an FC layer of 2-layer narrow NN provides lower accuracy than the default FC of ResNet and SpinalNet\footnote{https://github.com/dipuk0506/SpinalNet/tree/master/Jupyter\%20Notebooks /SpinalNet\_vs\_2L\_FC}. Therefore, we compare SpinalNet with the default FC of ResNet.

\subsubsection{CIFAR-10}
We train pre-trained VGG-19\_bn and Wide-Resnet-101 models to the CIFAR-10 data. We investigate both the traditional FC layers and Spinal FC layers. The SpinalNet achieves a significant improvement in the VGG network in terms of accuracy and the number of parameters. However, there is a slightly lower performance in the Wide-ResNet with the Spinal FC, compared to the original Wide-ResNet.

\subsubsection{CIFAR-100}
As usual, the SpinalNet performs well with the VGG network on the CIFAR-100 dataset. We receive a superior accuracy of 88.34\% with a wide Spinal FC of 512 neurons with dropout on the Wide-ResNet-101. It is one of the top ten reported results on the CIFAR-100 dataset. Results are presented in the second segment of table \ref{Perf_TL}.

\subsubsection{Caltech-101}
The SpinalNet performs well with both the VGG and the Wide-ResNet network on the Caltech-101 dataset compared to traditional fully connected layers. We receive SOTA performance for the Caltech-101 dataset.

\subsubsection{Bird225}
The SpinalNet performs well with both the VGG and the Wide-ResNet network on the Bird225 dataset. We receive SOTA performance for the Bird225 dataset. Results are available as the fourth segment of table \ref{Perf_TL}. However, the provider of the dataset has added more images to the dataset with new classes.

\subsubsection{Stanford Cars}
Although SpinalNet performs well with the VGG model, the accuracy with the Wide-ResNet model is equal. Moreover, our results are not one of the top 20 results in the Stanford Car leaderboard. The probable reason for the poor performance is the high resolution of the image. Several other researchers are concentrating on certain parts of the image to get high accuracy.

\subsubsection{SVHN}
Although we have received a slightly better result with the Spinal FC for both networks, our performance is not one of the top twenty performances. The probable reason for getting such poor performance is the training with a different kind of data. Most classes in the Imagenet dataset are not digits. Pre-training with different types of data may limit the performance.

\subsubsection{CINIC-10}
We receive significant improvements in terms of accuracy for both VGG and Wide-ResNet neural networks. Our result with Wide-ResNet Spinal FC is one of the top five reported results. Detailed procedures of data augmentation and simulations are available in the GitHub code.

\subsubsection{STL-10}
The SpinalNet performs well with both the VGG and the Wide-ResNet network on the STL-10 dataset. We receive SOTA performance for the STL-10 dataset.

\subsubsection{Oxford 102 Flower}
We trained pre-trained VGG-19\_bn and Wide-Resnet-101 to the Oxford 102 Flower data. We investigate both the traditional FC layers and Spinal FC layers. The SpinalNet achieves a significant improvement in the VGG network in terms of accuracy and the number of parameters. There is a slightly lower performance in the wide ResNet. Our results with the Wide ResNet networks are among the top 5 accuracies.

\subsubsection{Fruits 360}
Fruits 360 dataset is one of the easiest machine learning problems. Several programmers have released codes with more than 99.9\% efficiency in Kaggle for that dataset. We achieve promising performance with the SpinalNet on the Fruit 360 dataset with both the VGG and the Wide-ResNet networks. The Wide-ResNet-101 with Spinal FC provides 100\% accuracy. However, another paper has reported the same efficiency this year.

\subsubsection{Other Investigated Datasets}
We also investigate the Wide-ResNet101 with SpinalNet fully connected layers in `Intel Image Classification' and `10 Monkey Species' datasets in Kaggle and receive accuracies of 93.77\% and 99.26\% respectively.

\subsection{SOTA Performances}
Table \ref{SOTA_TAB} presents SOTA performances for our investigated datasets. Combining SpinalNet with VGG-5 provides near SOTA or SOTA performance in MNIST datasets. MNIST SOTA models are usually combined models and have the best performance for a specific dataset. We also obtain SOTA performance in five MNIST datasets. The prime reason for obtaining SOTA is not the proposed network. As these datasets are new, very few researchers investigated these data, and our result is the best among the reported results. We also apply transferred initialization with a spinal fully connected layer and achieve SOTA performance in several color image datasets.

\section{Prospects of SpinalNet}
There are many rooms for further investigation and improvement of SpinalNet. The rest of this section presents several prospects of SpinalNet.

\subsection{Auto Dimension Reduction}
Dimension reduction is a popular technique of reducing the number of inputs to a neural network without facing noticeable performance degradation \cite{vohra2019active}. The input combination of the NN network may contain a large number of inter-related or irrelevant data. As the proposed SpinalNet takes input in every layer and there are fewer neurons per hidden layer than the total number of inputs, the SpinalNet may automatically discard irrelevant data. Moreover, the necessity for dimension reduction may decrease, as a large number of input features do not increase the computation greatly.

\subsection{Very Deep NN}
The computation performed inside the proposed SpinalNet increases linearly with the increase in depth. The SpinalNet has outputs at every layer. Moreover, gradual training may enable us to increase SpinalNet depth gradually.

\subsection{Spinal Hidden Layer}
\begin{figure}
  \centering
  \includegraphics[width=3.2in,angle=0]{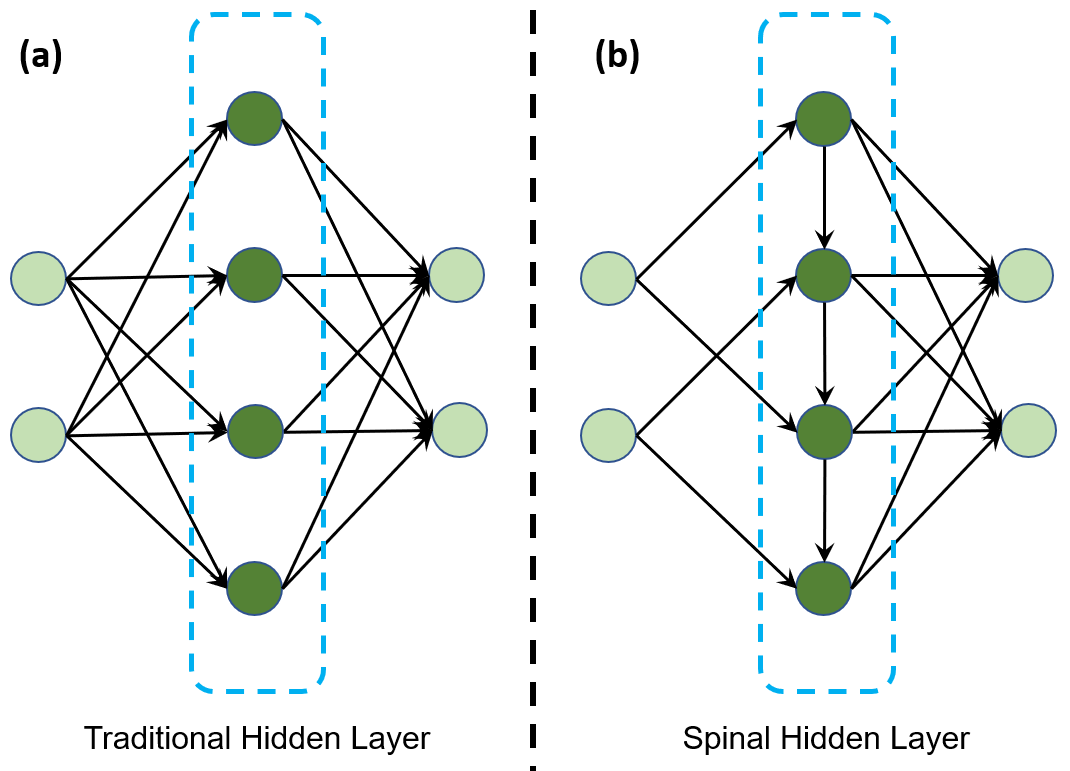}
  \caption{Any traditional hidden layer can be converted to a spinal hidden layer. The traditional hidden layer in (a) is converted to a spinal hidden layer in (b). A spinal hidden layer has the structure of the proposed SpinalNet.}
  \label{SpinalHL}
\end{figure}

This paper presents the SpinalNet as an independent network and as the fully connected layer of a CNN. The SpinalNet can also replace a wide hidden layer of a traditional NN. Fig. \ref{SpinalHL} presents how a SpinalNet can replace a traditional hidden layer. The figure shows two inputs and one neuron per sub-layer. The number of inputs can be large and the input can be segmented into more than two segments. Similarly, one sub-layer may hold more than one neuron.  The number of sub-layer can be more than or less than four.  

\subsection{Better Accuracy and New Datasets}
The SpinalNet may achieve higher accuracy with augmented datasets and its different structural variants. We apply SpinalNet as fully connected layers on several other networks to achieve higher accuracies. Researchers may apply SpinalNet for different datasets, new applications \cite{kabir2018neural,rahman2018unified, neven2018towards}, and combining with other networks in the future. 

\subsection{NN Ensemble and Voting}
Recently Mo Kweon et al. perform ensemble and voting from two different VGG networks and ResNet to achieve better performance \cite{VGG5}. Researchers may use different NNs within SpinalNet to get better performance.

After the online appearance of the paper, a researcher applied SpinalNet to predict DNA N6-Methyladenine Sites in Genomes \cite{abbas2020spinenet}. One researcher is also thinking of applying it for zero-shot learning \cite{pourpanah2020review}. Another recent ICLR submission applied SpinalNet along with other networks to explore supervised representation learning \cite{Anonymous2021ICLR}. Future research work may also deal with more scholarly analysis, such as sensitivity, adversarial learning, dropout uncertainty, etc \cite{pizarroso2020neuralsens, kamal2021alzheimer, theisen2021evaluating}. 

\section{Conclusion}
This paper presented a novel DNN model named SpinalNet. SpinalNet is built by mimicking the chordate nervous system, which has a unique way of connecting a large number of sensing information and taking local decisions. One major drawback of feed-forward NN models is their computational intensiveness for large inputs. Therefore, taking inputs gradually and considering local decisions similar to our spinal cord decreases computations. We also present the effectiveness of SpinalNet on several well-known benchmark datasets leading to the improvement of the classification accuracy and regression error. Moreover, the SpinalNet is usually less computation-intensive than its counterpart. Combining with VGG-5, SpinalNet has achieved SOTA performance in several handwritten character recognition datasets. With the transferred initialization, the SpinalNet has achieved SOTA performance in several color image datasets. Researchers may try to improve the accuracy of the proposed SpinalNet and apply the improved SpinalNet to a wide range of real-world scenarios.

\bibliographystyle{IEEEtran}
\bibliography{Ref}

\end{document}